# COVID-19 event extraction from Twitter via extractive question answering with continuous prompts

Yuhang JIANG[a] and Ramakanth KAVULURU[b,1]
[a] *Department of Computer Science, University of Kentucky, Lexington KY USA*
[b] *Division of Biomedical Informatics, University of Kentucky, Lexington KY USA*

**Abstract.** As COVID-19 ravages the world, social media analytics could augment traditional surveys in assessing how the pandemic evolves and capturing consumer chatter that could help healthcare agencies in addressing it. This typically involves mining disclosure events that mention testing positive for the disease or discussions surrounding perceptions and beliefs in preventative or treatment options. The 2020 shared task on COVID-19 event extraction (conducted as part of the W-NUT workshop during the EMNLP conference) introduced a new Twitter dataset for benchmarking event extraction from COVID-19 tweets. In this paper, we cast the problem of event extraction as extractive question answering using recent advances in continuous prompting in language models. On the shared task test dataset, our approach leads to over 5% absolute micro-averaged F1-score improvement over prior best results, across all COVID-19 event slots. Our ablation study shows that continuous prompts have a major impact on the eventual performance.

## 1. Introduction

Social media has emerged as a double-edged sword in the health communication world. Increasingly, consumers are seeking health information from online social platforms [1]. Hence, it is imperative that health agencies and professionals disseminate accurate information in a timely manner in appropriate social networks. On the other hand, misinformation is also proliferating on these platforms. The sudden rise of misinformation and the need to counter it has never been more urgent than during the ongoing COVID-19 pandemic [2]. For example, our prior efforts indicated concerted chatter surrounding promotions of nicotine, smoking, and vaping as potentially helpful to prevent/treat COVID-19, without any substantial evidence [3]. As smart phones become ubiquitous in the world, it's easy to both produce and consume information about health-related events. During the pandemic, these include disclosures of people testing positive or negative for COVID-19 and their takes/perceptions on what strategies or medications worked (or did not work) for them to prevent or treat the condition. If these are carefully extracted from massive online posts, they could help health agencies detect new spikes in infections and help track most frequently mentioned therapeutic options (some of which could be misinformation). The 2020 COVID-19 event extraction shared task[2] tackled this problem as part of the workshop on noisy user generated text (W-NUT) at the EMNLP 2020 conference. The organizers of the shared task created a manually curated dataset of five event types each with multiple slots (more in the next section). The overall aim was to facilitate a benchmark for COVID-19 event extraction on social media and to facilitate the creation of a massive COVID knowledge base of events that can be queried in a structured manner. In this paper, we improve on the prior best results [4] by over 5% in absolute micro-averaged F1 score on the test set of the shared task. To do this, we design a new approach using recent advances in continuous prompts for language models (LMs) to cast the event extraction task as an extractive question answering (QA) problem via questions prepended with the continuous prompts. Our code and config details are here: https://github.com/bionlproc/twitter-covid-QA-extraction

---
[1] Corresponding Author: Ramakanth Kavuluru (ramakanth.kavuluru@uky.edu).

## 2. Methods

### 2.1. The COVID-19 event extraction dataset

The five event types are (1) tested positive (TPos) (2) tested negative (TNeg) (3) cannot test (CT) (4) death (D) and (5) cure and prevention (C&P). The CT event was more relevant during the early phases of the pandemic when it was very difficult to find reliable and timely testing centers. Each event has its own slots that characterize the event. For example, a C&P event has three slots: (a) *what*: which method of cure/prevention is being mentioned? (b) *opinion*: does the author of the tweet believe that the cure/prevention is effective? (c) *who*: who is promoting the cure/prevention? The slot fillers are typically spans of text in the tweet that answer the corresponding slot question. We can see that (b) is binary response and hence not a span of text and (c) can have a special "author of the tweet" non-span label if the author of the tweet is promoting the cure and they don't explicitly refer to themselves in the tweet text. For example, consider the sentence: *"No doubt that **vaping** could have prevented a multitude of Covid19 deaths as reported by some **French scientists**"*. The *what* slot answer here would be the 4[th] token "vaping", the *opinion* slot's binary label would be YES (because the author of the tweet appears to believe in the effectiveness), and the *who* slot's answer is the last bigram "French scientists". For the TPos event, besides the expected *who*, *where*, and *when* slots, there are slots for *recent travel* and *employer* capturing their recent visit to a place before they tested positive and the company/org they work for, respectively. It's not hard to see that some slots do not necessarily have a valid span in the text. If recent travel and employer are not discussed in a tweet, the model is expected to not identify any spans. For a full description of all slots for all five events, please see the paper by the dataset creators [4]. The dataset has a total of 7500 training and development tweets and 2500 test set tweets annotated with various events.

### 2.2. Baselines and our methods

The main baseline is *bidirectional encoder representations from transformer*s (BERT), a well-known transformer model for NLP applications [5]. The current best results are from the shared task organizers Zong et al. [4] by using a special BERT model trained on COVID-19 tweets called the COVID-Twitter BERT (CT-BERT [6]) model (specifically, *covid-twitter-bert* on HuggingFace). We note that the candidate tweets for each event type are selected using special keyword-based queries as described in the appendix of Zong et al.'s paper [4]. The main approach used in all these prior approaches is to select possible different spans of an input tweet and predict if each of them answers the question corresponding to a particular slot. This can sometimes yield two different spans for the same slot, which is allowed in the dataset (e.g., for the "who" TPos slot, more than one person mentioned in the tweet could be testing positive). This approach can be very expensive as the number of candidate spans is typically all noun phrases and named entities mentioned in the tweet.

---
[2] http://noisy-text.github.io/2020/extract_covid19_event-shared_task.html

Our main approach is to map the slot filling task as an extractive QA task by passing the slot question text Q along with the tweet text T, separated by a special SEP token, as the input <Q> [SEP] <T> to a transformer language model (LM). It is trained with the output being the begin and end position tags for the tokens corresponding to the span that answers the question (for the slot being filled.) We begin with the well-known RoBERTa pretrained LM trained on the popular SQuAD QA dataset (V2) (specifically, *deepset/roberta-large-squad2* on HuggingFace). Since we are using a QA strategy, it is reasonable for us to use an LM trained on a QA dataset. We train and fine-tune this model using the training and development instances of the COVID event task. The parameter sizes of the state-of-the-art CT-BERT model and our fine-tuned RoBERTa model are both near 350 million and as such our model is not inherently more expensive to train. However, it is much faster at test time because we do not have to run a BERT classifier for each noun phrase; the model simply outputs the span boundaries of the answer.

We introduce a new variation besides passing the question text Q to the LM, inspired by the notion of continuous prompts. Although the question text Q encodes the information need, there are often multiple ways of asking the same question and it is well known that different variations of the same question may have different performances. So, to counter this, continuous prompts (additional trainable parameter vectors) were introduced recently by Li and Liang [7] as part of their so called "prefix tuning" strategy. The central idea is to use new "virtual tokens" that are not part of the base LM vocabulary as prefixes for downstream tasks. A high-level schematic of our approach is in Figure 1 with virtual tokens prepended to the input for each slot. Based on experiments with the development dataset, we determined 60 virtual tokens were apt for each slot, each with a dimension of 1024 generated through a multi-layer perceptron network that further parametrizes the model.

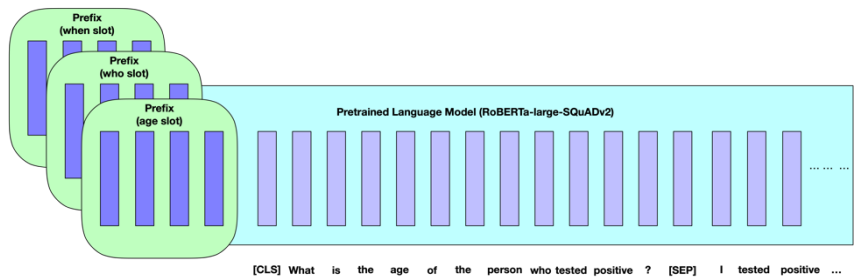

Figure 1: Our extractive QA architecture with continuous prompts for COVID event extraction

A few final considerations in our method involve some nuanced aspects of the dataset. We encode a missing slot with the output of the [CLS] (special start token standard in any LM) token span. For the "author of the tweet" outcome for *who* slot (e.g.,

"just tested positive folks"), at training time, we select the whole tweet as the gold outcome. At test time, if the model predicts a span that is longer than eight tokens, we output the "author of the tweet" label as it is inconceivable to have a *who* slot spanning many tokens. We transform the 3-way *gender* slot (male, female, unspecified) into two 2-way slots *gender-male* (male or not) and *gender-female* (female or not) to account for our span-based method. Because the RoBERTa model only outputs a single span and given our task allows for multiple spans per slot, we split the single span based on commas and conjunctions to identify potential multiple spans. Finally, all shared task participants and the annotators who created the dataset were only allowed to select answer spans from pre-chunked noun phrases and named entities. These chunks are provided for each tweet as part of the dataset. The baseline methods and CT-BERT operate on these chunks; since ours is a span detection method, our spans do not always match with organizer provided chunks. We address this by identifying the closest match of a predicted span using Jaccard similarity among tweet chunks. If there is no overlap at all with a chunk, we predict a missing slot. Our model is fine-tuned with a learning rate $4 \times 10^{-6}$ using the Adam optimizer for 8 epochs.

## 3. Results

We first present the event level classification results for BERT, CT-BERT, and our approach. We note that since different search terms were used by the dataset creators for different types, event type classification boils down to the binary setting where an event is identified if any of the slots returns a valid non-[CLS] span. From Table 1, we clearly see that across all event types our method outperforms the prior best results [4] by double-digit margins in terms of F1-scores. The macro-average F1 score is 89.1 for our model compared to 77.4 for CT-BERT. While this is encouraging, identifying the event type is not as useful unless individual slots are accurately filled for those events.

Table 1: F1 scores for event level classification

|  | **BERT** | **CT-BERT** | **Ours** |
|---|---|---|---|
| **Tested Positive** | 90.0 | 88.5 | **94.5** |
| **Tested Negative** | 72.0 | 77.2 | **89.0** |
| **Cannot Test** | 72.0 | 72.8 | **86.6** |
| **Death** | 73.0 | 78.7 | **88.0** |
| **Cure and Prev.** | 64.0 | 69.8 | **87.9** |

*Overall, the micro-averaged F1 score across all 31 slots (across the five event types) for CT-BERT is 67.0 and for our approach is 72.5, indicating over 5% absolute improvement.* All individual slot level scores are difficult to display within space constraints. However, a common trend we observed is that our approach greatly improves recall compared to the CT-BERT method. However, it sometimes loses some precision but, overall, we see more balance between precision and recall with our method. We demonstrate this via slot-level performances for the *cure and prevention* event shown in Table 2. With the bold values we can see that CT-BERT is better in precision, while our method is superior in recall and F1-score. We also conducted ablation experiments and found that our micro-averaged F1 across all slots drops from 72.5 to 66.8 if we drop

the continuous prompts and simply used the **Q [SEP] T** input to the model, indicating major importance of those virtual tokens. Likewise, we see the score dip to 67.9 if we don't use pre-training on the SQuAD dataset, confirming that pretraining on a task that is similar to the downstream task is very important.

Table 2: Precision, recall, F1 scores for the Cure and Prevention event type

| **Cure & Prev slots** | **CT-BERT** | | | **Ours** | | |
|---|---|---|---|---|---|---|
| | P | R | F1 | P | R | F1 |
| **Opinion** | 85 | 59 | 69 | 68 | 82 | 74 |
| **What** | 83 | 64 | 72 | 74 | 70 | 72 |
| **Who** | 87 | 37 | 51 | 75 | 59 | 66 |

## 4. Discussion and Conclusion

From Tables 1 and 2, and the overall micro-average improvement of over 5% in slot filling for the full test set compared with prior best results that use CT-BERT [4], we believe our method is a nontrivial advance in pandemic related event extraction from Twitter data. Though our focus is exclusively on COVID-19, we believe the general event types and associated slots are applicable to any pandemic situation, in general. We conducted preliminary error analyses and noticed that we need to improve further in mapping our spans to chunks provided by the dataset, especially for the age slot where the Jaccard similarity is not very reliable. Additionally, our analyses revealed ambiguous/subtle cases that could also be construed as correct extractions. For instance, our model extracts Notre Dame for the *employer* slot from the tweet "A Notre Dame football player has tested positive for COVID-19". Although the tweet does not clearly say the player works for the university, it satisfies the general intention behind the *employer* slot. Despites these limitations, we believe our model will lead to improved social media based infodemiology studies for pandemics.

**Acknowledgments**: This work is supported by the U.S. National Library of Medicine through grant R01LM013240.